\documentclass[cameraready]{Interspeech}
\usepackage{multicol}
\usepackage{multirow}
\usepackage{arydshln}
\usepackage{hyperref}

\title{Bagpiper-TTS: Natural Language Guided Universal Speech Synthesis}

\author[affiliation={1}]{Jinchuan}{Tian}
\author[affiliation={1}]{Haoran}{Wang}
\author[affiliation={1}]{Siddhant}{Arora}
\author[affiliation={2}]{Takashi}{Maekaku}
\author[affiliation={2}]{Keita}{Goto}
\author[affiliation={2}]{Jin}{Sakuma}
\author[affiliation={2}]{Yusuke}{Shinohara}
\author[affiliation={3}]{Chao-Han Huck}{Yang}
\author[affiliation={1}]{Shinji}{Watanabe}

\address{
    $^1$ Language Technologies Institute, Carnegie Mellon University \\
    $^2$ LY Corporation \quad
    $^3$ NVIDIA Research
}

\email{jinchuat@andrew.cmu.edu}

\keywords{Speech Language Model, Text-to-Speech, Rich Caption}

\usepackage{comment}

\begin{document}

\maketitle

\begin{abstract}
    Classical TTS systems typically rely on rigid input formats and predefined metadata slots, limiting their ability to fulfill flexible user requirements. This paper introduces Bagpiper-TTS, a universal speech synthesis system that deals with diverse natural language user requests. 
    Given a natural language prompt, Bagpiper-TTS first reasons over the user's intent to derive a \textit{rich caption}, i.e., a comprehensive textual blueprint encompassing both transcription and nuanced metadata. Subsequently, this caption guides the synthesis of the target speech. Our model inherently supports a broad spectrum of tasks besides classical TTS applications, including multi-talker, intent-to-speech, role-play synthesis, singing voice synthesis, and more. Experimental results demonstrate that Bagpiper-TTS achieves an $1.7\%$ Word Error Rate (WER) on the Seed-TTS-Eval benchmark and match the performance of dedicated models in both LLM-as-a-judge and human subjective evaluations across multiple applications.\footnote{Demo, code, data, and checkpoints are available at our \href{https://bagpipertts.github.io/bagpiper_tts_demo/}{\textcolor{blue}{ HomePage}}.}
\end{abstract}

\section{Introduction}
Traditional text-to-speech (TTS) systems are primarily engineered to synthesize speech from text transcriptions, often augmented by specific metadata such as speaker identity, language, emotion, and stress \cite{tan2021survey, xie2025towards}. To incorporate these attributes, early designs rely on specialized, disjointed modules (e.g., dedicated speaker encoders) to capture paralinguistic information \cite{fastspeech, tacotron}. While recent advancements in large language models (LLMs) \cite{wang2023neural, peng2026vibevoice, tian2025opuslm, maiti2024voxtlm, yang2025simplespeech, hu2026qwen3}, diffusion-based \cite{le2023voicebox, ju2024naturalspeech, chen2025f5, gao2023e3} models, and their hybrid \cite{du2025cosyvoice, anastassiou2024seed} have significantly enhanced fidelity and naturalness, they largely inherit this rigid input paradigm. These models typically require structured, predefined metadata slot filling, which creates a fundamental mismatch with real-world applications where user requests are inherently fluid, unpredictable, and highly varied.

Furthermore, as the scope of TTS expands to include complex tasks such as multi-talker dialogue \cite{ju2025mooncast, peng2026vibevoice} and immersive role-play \cite{zhang2025omnicharacter}, the demand for intricate control over speech attributes has grown considerably. Current architectures struggle to integrate these diverse functionalities into a single, cohesive framework without sacrificing flexibility or increasing pipeline complexity. To bridge this gap, we argue for a shift toward a natural language interface capable of interpreting free-form user requests through a human-centric approach. Such an interface allows TTS systems to move beyond the constraints of fixed metadata slots, instead adapting dynamically to the nuanced requirements of any user-defined scenarios.

\begin{figure}
    \centering
    \includegraphics[width=\linewidth]{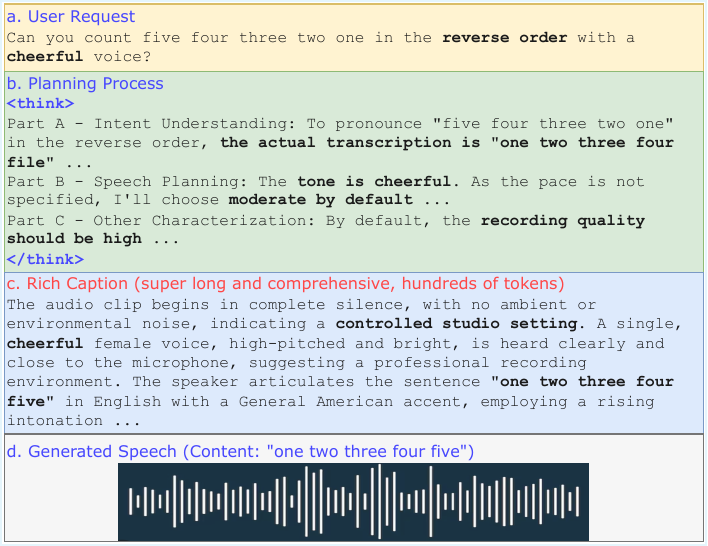}
    \caption{The \textit{Planning-Caption-Generation} Workflow of Bagpiper-TTS to address general-purpose text-to-speech request. Content is abbreviated.}
    \label{fig:workflow}
    \vspace{-20pt}
\end{figure}

This paper presents Bagpiper-TTS, a framework designed to resolve the limitations of traditional, rigid TTS systems by mediating free-form user requests through rich captions \cite{ma2025omni, anonymous2025videosalmonn, fiscus2006rich, chen2026audiochat}. As shown in Fig.\ref{fig:workflow}.c, these captions serve as a comprehensive textual blueprint, meticulously encoding both the transcription and a wide spectrum of paralinguistic metadata. 
Given the user request, the model first interprets the user's intent through a planning process within the text domain, subsequently synthesizing a highly detailed rich caption that acts as a generation blueprint. By leveraging captions that can scale up to hundreds of tokens for a 30-second speech clip, the pipeline theoretically accommodates any arbitrary user inputs, translating diverse natural language instructions into high-fidelity speech.

Bagpiper-TTS is built upon existing Bagpiper-Base \cite{tian2026bagpiper}, an audio foundational model pre-trained extensively on large-scale rich-caption-to-speech tasks with strong text planning capability preserved. As rich captions contain comprehensive metadata for speech, such pre-training strongly aligns the natural language and speech attributes and facilitate downstream fine-tuning.
To unlock the model's capacity for free-form instruction following, we curated a specialized fine-tuning corpus using a sophisticated simulation pipeline. Starting with high-quality caption-speech pairs (Fig.\ref{fig:workflow}.c, d), we utilized large language models (LLMs) to reverse-engineer potential user requests (Fig.\ref{fig:workflow}.a). We then constructed a text-based planning process to bridge the gap between initial requests and final rich captions (Fig.\ref{fig:workflow}.b). To ensure the highest data integrity, the resulting triplets (i.e., user requests, planning process, and rich captions) were rigorously audited by an LLM for logical coherence, while Gemini-3-Flash is optionally employed as a multimodal validator to verify the consistency between the synthesized captions and the ground-truth speech.
Through this paradigm, Bagpiper-TTS achieves robust capabilities across classical, multi-talker, intent-to-speech, role-play synthesis, and singing voice synthesis, all using natural language as the universal portal. We further augmented the training set with "general-purpose" data to simulate the highly diverse and non-standard requests encountered in real-world scenarios. For the classical TTS application, we validated our approach on the Seed-TTS-Eval (En) benchmark, where Bagpiper-TTS achieved a Word Error Rate (WER) of $1.7\%$ on the classical TTS task. 
Extensive comparative studies on other advanced TTS applications demonstrate that our model successfully fulfills the free-form user requests, as confirmed by both LLM-as-a-judge metrics and human subjective evaluations.

\section{Bagpiper-TTS}

\subsection{Bagpiper-Base} \label{sec:arch}
Bagpiper-TTS is built upon existing Bagpiper-Base \cite{tian2026bagpiper}, an audio-centric foundational model designed for cross-modal understanding and synthesis. Its architecture utilizes the Qwen3-8B-Base \cite{yang2025qwen3} decoder-only LLM as its computational backbone, which provides strong text-only capability. To support audio generation, it tokenizes the audio with the multi-stream X-Codec \cite{xcodec} operating at 50Hz, producing 8 discrete codes per frame as the audio predicting targets.

Bagpiper-Base was pre-trained on 600 billion tokens, including caption-to-audio synthesis, audio-to-caption understanding, and pure text modeling. Its pre-training spans speech, music, and environmental sounds, so it can support universal audio generation in the downstream. This large-scale pre-training ensures a robust alignment between detailed textual descriptions and their corresponding acoustic realizations while preserving the backbone's native text-based reasoning capabilities. We refer readers to \cite{tian2026bagpiper} for extensive architectural and pre-training details.

\subsection{Natural Language as a Universal Interface}\label{sec:workflow}
In contrast to traditional systems that rely on rigid, slot-based input formats, Bagpiper-TTS adopts natural language as a universal interface. This approach dynamically encodes transcriptions and standard metadata, e.g., speaker identity and prosody, alongside complex contextual nuances, including role-play character traits and atmospheric background elements. By providing a human-centric interface, Bagpiper-TTS empowers users to combine instructions using arbitrary linguistic styles and structures, effectively bridging the gap between user intent and acoustic execution. Consequently, the system democratizes sophisticated speech synthesis, shifting the requirement away from technical expertise toward basic natural language proficiency, making high-quality TTS accessible to non-expert users.

\noindent\textbf{Planning-Caption-Generation Workflow:}
To process flexible natural language requests, Bagpiper-TTS follows a three-stage hierarchical workflow (see Fig.~\ref{fig:workflow}) that is executed entirely in an end-to-end manner within a single unified model.
\textbf{Textual Planning:} The model performs an initial reasoning step to comprehend user intent, outlining a conceptual summary that includes transcription layouts, speech events, and stylistic constraints.
\textbf{Rich Caption Synthesis:} Based on the planning output, the model generates a comprehensive \textit{rich caption}, a dense textual blueprint that formalizes the paralinguistic and acoustic requirements.
\textbf{Speech Generation:} The rich caption serves as the direct guidance for speech synthesis. This stage leverages the strong caption-to-speech alignment established during the Bagpiper-Base pre-training phase, ensuring high-fidelity adherence to the user's original request.

\subsection{Fine-Tuning Data Simulation}

\subsubsection{Data Simulation Pipeline} \label{sec:simulation}
\begin{figure}
    \centering
    \includegraphics[width=0.9\linewidth]{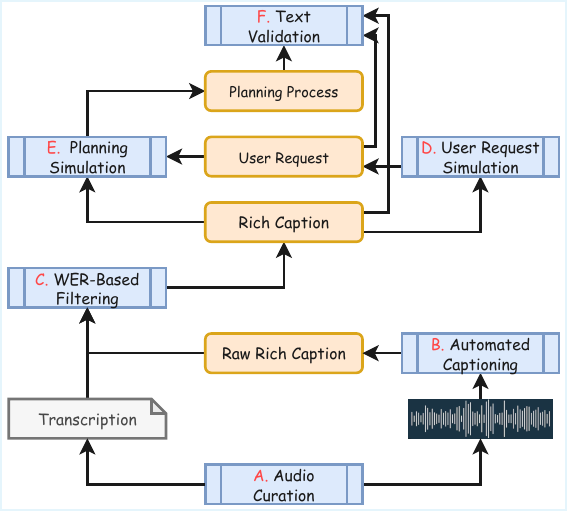}
    \caption{Flowchart of fine-tuning data simulation pipeline}
    \label{fig:pipeline}
    \vspace{-20pt}
\end{figure}

Our data simulation pipeline is illustrated in Fig.~\ref{fig:pipeline}\footnote{Unless otherwise specified, we consistently adopt Qwen3-235B-A22B-Instruct-FP8 as the primary text LLM for all data processing tasks.}. The pipeline follows a six-step procedure designed to generate free-form user requests with corresponding model responses:

\textbf{Step A: Speech Curation.} 
For each speech synthesis application described in \S\ref{sec:application}, we first curate a set of proper speech clips, typically accompanied by ground-truth transcriptions to serve as an anchor for subsequent stages.\footnote{We obtain pseudo-transcriptions with Qwen3-ASR \cite{shi2026qwen3} if no transcription provided.}
\textbf{Step B: Automated Captioning.} We process the speech clips through an automatic captioning model, Qwen-30B-A3B-Captioner \cite{qwen3omni}, to generate comprehensive rich captions. We acknowledge that captions may contain hallucinations regarding various acoustic or linguistic attributes.
\textbf{Step C: WER-Based Filtering.} To ensure the correctness of transcription in rich caption, we extract the transcription from the rich caption using the text LLM and calculate the Word Error Rate (WER) against the ground-truth. Examples exceeding a specific WER threshold are discarded (e.g., 0\% to enforce exact match for most applications). Optionally, the LLM is prompted to refine the rich caption to align with the ground-truth transcription, thereby maximizing data retention while ensuring accuracy.

\textbf{Step D: User Request Simulation.} We reverse-engineer diverse user requests by prompting the text LLM with the rich captions. To maximize variance, we instruct the LLM to vary the request length, linguistic style, and the ordering of metadata components (e.g., speaker identity vs. transcription), tailored to each specific TTS application.
\textbf{Step E: Planning Process Simulation.} We utilize the text LLM to construct a textual planning process that bridges the logical gap between raw user requests and structured rich captions. This planning step systematically addresses three parts, as in Fig.\ref{fig:workflow}.b: (1) interpreting user intent and provided metadata; (2) organizing speech-specific attributes such as speaker identity and pacing; and (3) incorporating non-speech elements like ambient sounds and recording conditions.

\textbf{Step F: Textual Consistency Validation.} We validate the logical coherence among the generated triplets, i.e., user requests, planning processes, and rich captions. Using an LLM-as-a-judge approach, each sample is evaluated on a scale of 1–5 across multiple task-specific metrics (e.g., does the audio described in rich caption fulfill the user request?). The LLM is instructed to be very picky and selective, but be tolerant of the overly comprehensive rich captions. We retain only those samples with an average score above 3.5 and no individual metric score below 3.

\subsubsection{Target Applications} \label{sec:application}
\begin{figure}
    \centering
    \includegraphics[width=\linewidth]{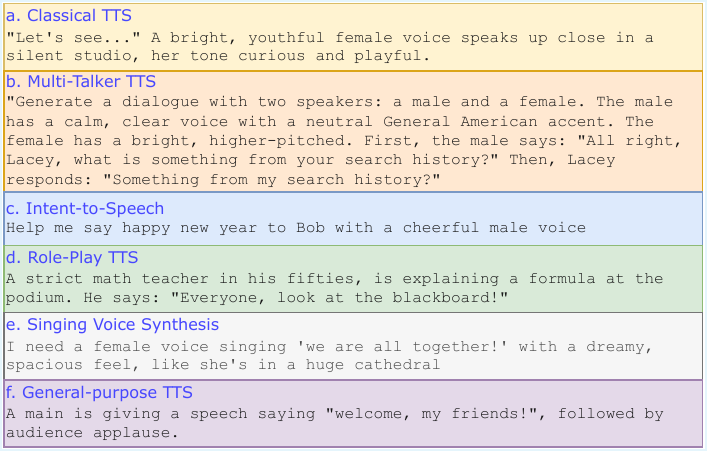}
    \caption{Example user requests for each TTS application.}
    \label{fig:application}
    \vspace{-20pt}
\end{figure}
Leveraging the simulation framework described in \S\ref{sec:workflow}, we curate fine-tuning datasets for a diverse suite of applications, which mimics the free-form and flexible user requests. We further incorporate a "general-purpose" subset to address the out-of-definition user requests. Due to space constraints, we outline the high-level design considerations for each application below; comprehensive implementation details are available in our open-source codebase. Fig.\ref{fig:application} presents examples for each of these applications.

\noindent\textbf{Classical TTS:}
This task represents the standard scenario where users provide explicit transcriptions alongside direct stylistic metadata (e.g., specific emotional labels) \cite{yang2024instructtts}. We utilize LibriTTS-R \cite{koizumi2023libritts} for neutral, prosodically consistent speech and the Genshin \cite{simon3000_genshin_voice} and Starrail \cite{simon3000_starrail_voice} datasets for expressive, spontaneous speech. During user request simulation, metadata is selectively extracted from rich captions to mimic varied levels of user specification.

\noindent\textbf{Multi-talker TTS:}
This application generates multi-speaker speech, primarily in the form of spoken dialogues \cite{peng2026vibevoice, ju2025mooncast}. To curate suitable training speech, we merge segments from long-form recordings in Gigaspeech \cite{chen21o_interspeech} and SSSD \cite{sheikh25_interspeech}, and utilize VibeVoice-ASR \cite{peng2026vibevoice-asr} to capture multi-speaker intersections. In the subsequent simulation stages, we prioritize the distinct acoustic characterization of each speaker and the accuracy of their temporal interleaving within each simulated textual component.

\noindent\textbf{Intent-to-speech:}
In this novel setup, the exact transcription is absent from the user request and must be inferred from the context (e.g., user prompts \textit{"Wish Bob a happy new year in a cheerful female voice"} and the system produces \textit{"Hi, Bob, Happy New Year!"}). We prompt the text LLM to curate speech segments in Bagpiper-Base pre-training that express clear intentions and are of high communication value. During the planning phase, we ensure the transcription is intentionally excluded from the user request but correctly inferred within the planning process to maintain alignment with the synthesized speech.

\noindent\textbf{Role-play TTS:}
Role-play TTS requires the model to infer acoustic features indirectly from character descriptions \cite{zhang2025omnicharacter, hu2026qwen3}.\footnote{This is similar to the \textit{voice design} feature in concurrent work \cite{hu2026qwen3}.} For instance, a character described as a \textit{"senior mathematics professor"} may imply a deep, authoritative tone. 
We curate the expressive examples in Bagpiper-Base pre-training and reverse-engineer the character descriptions by text LLM based on rich captions. Validation focuses on the logical consistency between the character description and the inferred acoustic realized in the rich caption.

\noindent\textbf{Singing Voice Synthesis (SVS):}
Unlike speech synthesis, SVS requires the model to generate melodic content, often accompanied by background music. 
Utilizing the singing samples in Bagpiper-Base pre-training, we obtain the pseudo transcription via Qwen3-ASR \cite{shi2026qwen3} and relax the WER constraints to 10\% during step C to account for melodic variance. 
We further enforce the inclusion of descriptive cues for background accompaniment across all triplet components to ensure harmonic consistency.

\noindent\textbf{General-purpose TTS:}
To accommodate requests that fall outside the task definition above, we randomly select 2M speech samples from the Bagpiper-Base pre-training data to create the \textit{general-purpose} subset, 
We utilize Claude 4.6 Opus to produce 40 distinctive application scenarios and ask the text LLM to pick applicable ones based on the rich captions during step D, and simulate and validate based on that scenario in steps E and F.

\begin{figure}[h]
    \centering
    \vspace{-10pt}
    \includegraphics[width=0.8\linewidth]{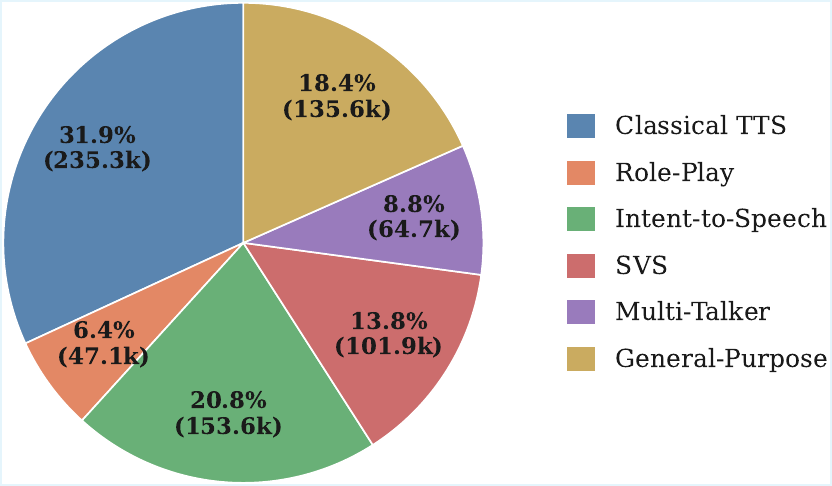}
    \caption{Fine-tuning data distribution for all applications}
    \label{fig:distribution}
    \vspace{-20pt}
\end{figure}

Our data does not include any system prompt, so the exact pre-defined application is agnostic to the model. The model comprehends and reacts to natural language user requests based on the requests only. 
Overall, we curate 738k examples in total. The distribution is demonstrated in Fig.\ref{fig:distribution}.

\subsection{Evaluation Protocols}
We evaluate the performance of Bagpiper-TTS across three distinct tiers: classical benchmarks, specialized applications, and general-purpose flexibility. For Classical TTS, we utilize the Seed-TTS-Eval (En) benchmark \cite{anastassiou2024seed}, prompting the system to generate speech in a plain voice.\footnote{We do not test the speaker similarity as our system does not accept audio prompts (e.g., reference speaker clip), as discussed in \S\ref{sec:limitation}.}

Evaluating Multi-talker, Intent-to-speech, Role-play, and Singing Voice Synthesis presents unique challenges due to a lack of standardized benchmarks. To address this, we simulate 300 unique user requests for each application using GPT-OSS-120B \cite{agarwal2025gpt}. To ensure the validity of our results, the test request simulation is strictly isolated from the fine-tuning data generation pipeline. Performance is quantified using the Word Error Rate (WER) alongside an LLM-as-a-judge framework, where Gemini-3-Flash scores task fulfillment on a scale of 1–5. 
Complementing these metrics, we conduct a subjective evaluation via Amazon Mechanical Turk (MTurk). 10 examples per application (derived from the full pipeline including Gemini-based filtering) are rated by at least three human annotators on a scale of 1–5, with 3 being acceptable. The sample selection for subjective evaluation is totally random with no cherry-pick.

Finally, for the General-purpose TTS tier, we forgo quantitative metrics in favor of qualitative analysis. Researchers manually prompt the system with hand-written, "out-of-definition" requests to test the model's reasoning capabilities, with detailed observations provided in \S\ref{sec:observation}.

\section{Experiments}
\subsection{Experimental Setup}
Building upon the Bagpiper-Base foundational model, we perform supervised fine-tuning on all our curated simulation datasets for 2 epochs to build this universal speech generation model. We utilize a global batch size of 160k tokens and maintain a constant learning rate of $1 \times 10^{-5}$.
During inference, we employ a decoupled Top-$k$ sampling strategy for both text and speech modalities, with specific configurations detailed in \cite{tian2026bagpiper}. Notably, we apply Classifier-Free Guidance (CFG) \cite{cfg} with a guidance scale of $\lambda=3$ during speech generation to enhance stylistic fidelity. The fine-tuning process was completed in 16 hours using 8 NVIDIA H100 GPUs.

\subsection{Results}
We first evaluate classical TTS performance, as summarized in Table~\ref{tab:classic}. The results indicate that, despite the added complexity of natural language prompts (where metadata and transcriptions are dynamically interleaved), our Bagpiper-TTS generates highly intelligible speech that is competitive with current state-of-the-art frontier models.

\begin{table}[h]
    \centering
    \caption{Classic TTS evaluation results on Seed-TTS-Eval (En). Bagpiper-TTS (ours) accepts natural language prompts.}
    \vspace{-12pt}
    \scalebox{0.8}{
    \begin{tabular}{lcc} \\
    \toprule
                    & WER \\
        \midrule
        CosyVoice 2 \cite{du2024cosyvoice}&  2.6\\
        VibeVoice   \cite{peng2026vibevoice} & 3.0  \\
        Qwen3-TTS   \cite{hu2026qwen3} & \textbf{1.5} & \\
        \midrule 
        Ours & 1.7 \\
        \bottomrule
    \end{tabular}}
    \vspace{-10pt}
    \label{tab:classic}
\end{table}

Evaluation results for advanced applications (\S\ref{sec:application}), including multi-talker dialogue, intent-to-speech, role-play, and SVS, are presented in Table~\ref{tab:advance}. These findings confirm that Bagpiper-TTS successfully fulfills a wide range of sophisticated synthesis tasks through the universal natural language portal, effectively addressing scenarios that remained underexplored in prior modular research. Across all categories, the model maintains a consistently lower Word Error Rate (WER) than baselines, establishing a robust performance baseline for speech intelligibility.

In our LLM-as-a-judge evaluation, Bagpiper-TTS achieved a mean fulfillment score of 4.09 across the four advanced applications, which suggests the model fulfills all four applications effectively in generally \textit{good} level.
The task fulfillment is additionally confirmed by the human-as-a-judge subjective evaluation, where our Bagpipier-TTS achieves an average score of 3.69, also proving that the proposed tasks are effectively addressed in this early exploration. Although there is a slight performance gap between our model and baselines (VibeVoice \cite{peng2026vibevoice} and YuE \cite{yuan2025yuescalingopenfoundation}), we note our model is a generalist model while others are more specialized. 

\begin{table}[t]
    \centering
    \caption{Evaluation results on advanced TTS applications (\S\ref{sec:application}). Bagpiper-TTS is a \textbf{single universal model that addresses all these applications} in an application-agnostic style. TF: Task Fulfillment score judged by Gemini-3-Flash.}
    \vspace{-5pt}
    \scalebox{0.8}{
    \begin{tabular}{llccccc}
    \toprule
                  &&  \multicolumn{2}{c}{Objective} & Subjective \\
         Application & Model & WER & TF & MOS \\
         \hline
         \multirow{2}{*}{Multi-Talker} & VibeVoice-1.5B \cite{peng2026vibevoice} & 4.6 & 3.32 & \bf{3.77} \\
                                       & Ours & \bf{4.2} & \bf{4.23} & 3.60 \\
         \hline
         Intent-to-Speech & Ours& - & 3.80 & 3.57 \\
         \hline 
         Role-Play TTS &  Ours & 2.0 & 3.72 & 3.93 \\
         \hline
         \multirow{2}{*}{SVS} & YuE \cite{yuan2025yuescalingopenfoundation} & 11.0 & 3.75 & \bf{3.73} \\
                                       & Ours & \bf{7.2} & \bf{4.60} & 3.67 \\
        \bottomrule
    \end{tabular}}
    \label{tab:advance}
    \vspace{-10pt}
\end{table}

\subsection{Qualitative Study} \label{sec:observation}
To demonstrate its versatility, we evaluate Bagpiper-TTS against complex, free-form requests that exceed the capabilities of existing modular systems. We observe that the model effectively interprets non-straightforward logic; for instance, when prompted to \textit{"count from one to five"} or \textit{"read five four three two one in reverse order,"} it correctly reasons through the instruction to synthesize the sequence \textit{"one two three four five"}. Furthermore, the model demonstrates sophisticated multimodal alignment by adapting both its acoustic delivery (e.g., a soft, restrained tone) and its linguistic wording to match nuanced requests like \textit{"gentle criticism"}, where the criticism is expressed, but the tone and wording are both polite and decent.
Beyond the final speech, the model’s internal planning—producing autonomous justifications such as \textit{"the content is for children, so the delivery should be warm"}—marks a clear departure from static slot-filling. This ability to resolve ambiguous metadata through latent judgment makes the system significantly more robust to unpredictable human instructions.
We strongly encourage readers to listen to our demo examples on our \href{https://bagpipertts.github.io/bagpiper_tts_demo/}{\textcolor{blue}{Bagpiper-TTS HomePage}}.

\section{Limitations} \label{sec:limitation}
Although we implement rigorous filtering, hallucinations persist across various stages of data simulation and model inference, particularly within the rich captions produced by the automated captioner. Furthermore, while this work focuses on textual natural language as a universal interface, we acknowledge that certain metadata are more effectively represented through acoustic grounding (e.g., using a reference audio prompt to define a speaker’s voice characteristics). 

\section{Conclusion}
This paper introduces Bagpiper-TTS, a framework that replaces rigid metadata inputs with a natural language interface for flexible, free-form speech synthesis. By employing an end-to-end reasoning workflow with rich captions, Bagpiper-TTS successfully interprets complex user intent across diverse tasks. Our results demonstrate that this human-centric approach achieves competitive performance on standard benchmarks while providing unprecedented versatility, marking a significant step toward intuitive, universal speech foundational models.

\newpage


\section{Use of Generative AI Disclosure}
Generative AI tools were utilized in the preparation of this manuscript primarily for stylistic polishing and grammatical refinement; they were not used to author significant original technical content. Additionally, as detailed in \S\ref{sec:workflow} and \S\ref{sec:simulation}, Large Language Models (LLMs) and multimodal foundational models were systematically employed within our data simulation pipeline to generate synthetic fine-tuning datasets.

\section{Acknowledgment}
Parts of this work used the PSC Bridges2 system and Delta/DeltaAI system at NCSA through allocations CIS210014 and IRI120008P from the ACCESS program, supported by NSF grants \#2138259,\#2138286, \#2138307, \#2137603, and \#2138296.

\bibliographystyle{IEEEtran}
\bibliography{mybib}

\end{document}